\def\year{2021}\relax
\documentclass[letterpaper]{article} 
\usepackage{aaai21}  
\usepackage{times}  
\usepackage{helvet} 
\usepackage{courier}  
\usepackage[hyphens]{url}  
\usepackage{graphicx} 

\usepackage{subfig}

\usepackage{tabularx}
\usepackage{booktabs}

\usepackage{natbib}  
\usepackage{caption} 
\title{Interactive Hierarchical Guidance using Language}
\author{
    Bharat Prakash\textsuperscript{\rm 1},
    Nicholas Waytowich\textsuperscript{\rm 2},
    Tim Oates\textsuperscript{\rm 1},
    Tinoosh Mohsenin\textsuperscript{\rm 1}, 
    \\
}
\affiliations{
    \textsuperscript{\rm 1}University of Maryland, Baltimore County,
    \textsuperscript{\rm 2}US Army Research Lab\\

    bhp1@umbc.edu\textsuperscript{\rm 1}, nicholas.r.waytowich.civ@mail.mil\textsuperscript{\rm 2},
    oates@cs.umbc.edu\textsuperscript{\rm 1}, tinoosh@umbc.edu\textsuperscript{\rm 1}

}

\begin{document}

\maketitle

\begin{abstract}
Reinforcement learning has been successful in many tasks ranging from robotic control, games, energy management etc. In complex real world environments with sparse rewards and long task horizons, sample efficiency is still a major challenge. Most complex tasks can be easily decomposed into high-level planning and low level control. Therefore, it is important to enable agents to leverage the hierarchical structure and decompose bigger tasks into multiple smaller sub-tasks. We introduce an approach where we use language to specify sub-tasks and a high-level planner issues language commands to a low level controller. The low-level controller executes the sub-tasks based on the language commands. Our experiments show that this method is able to solve complex long horizon planning tasks with limited human supervision. Using language has added benefit of interpretability and ability for expert humans to take over the high-level planning task and provide language commands if necessary.

\end{abstract}

\section{Introduction}

As autonomous agents get more advanced and capable we are more likely to see them operating along with humans in the same environment. In order to achieve efficient collaboration, both humans and autonomous agents need to understand each others intent. Although there are many non verbal ways to communicate, language is an effective widely used mode of communication and certainly has many advantages. Therefore, it is important to design autonomous agents which can understand and respond using natural language. Language has many inherent benefits; it can be used as abstractions to communicate plans/goals and or behaviours. It has a compositional structure to communicate complex information. It is a natural way for humans to communicate and we already have a lot of infrastructure that makes use of language in the real world e.g., instruction manuals, warning signs, navigation tools etc. Moreover mapping agent behavior with language is very useful for interpretability and understanding what the agents are `thinking' when making decisions. 

\begin{figure}
		\centering{\includegraphics[width=0.45\textwidth]{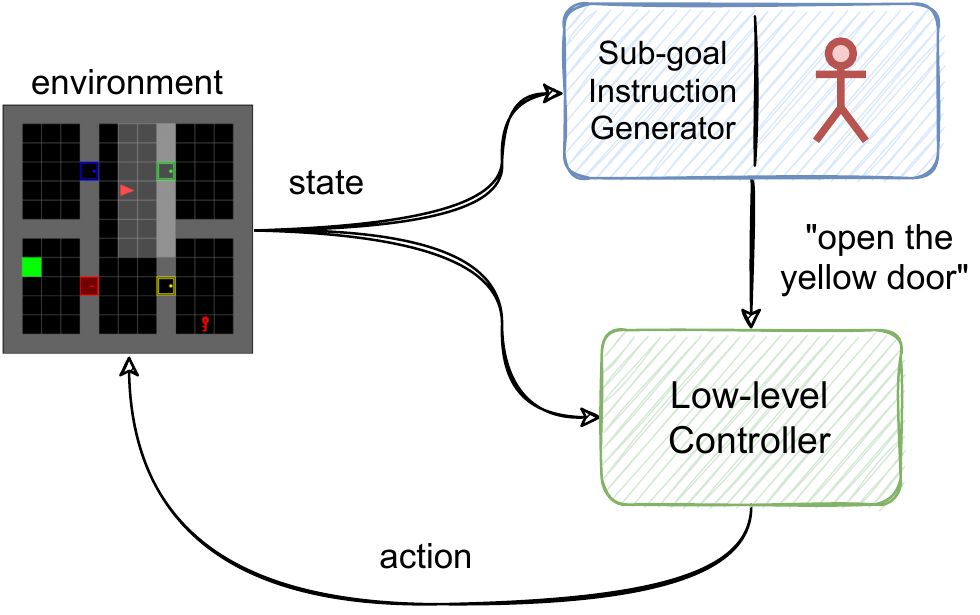}}
		\caption{The architecture consists of the sub-goal instruction generator which output sub-goals in the form of language. The low level controller is a language conditioned policy which accepts the current state and language command as input and outputs a series of actions which is executed in the environment.}
		\label{fig:hrl_arch}
\end{figure}

In this framework we propose to use language to specify high-level sub-tasks. Given a high-level objective and access to sub-tasks to complete a longer and complicated task, we can collect data where these entities are labeled using language descriptions. Then, the task is to learn a mapping between current state to sub-tasks in terms of language. In addition to that, we learn a mapping between sub-tasks to low-level agent behavior or actions. In the rest of the paper, we will use sub-tasks and sub-goals interchangeably.

\begin{figure*}
  \centering
  \subfloat{\includegraphics[scale=0.80]{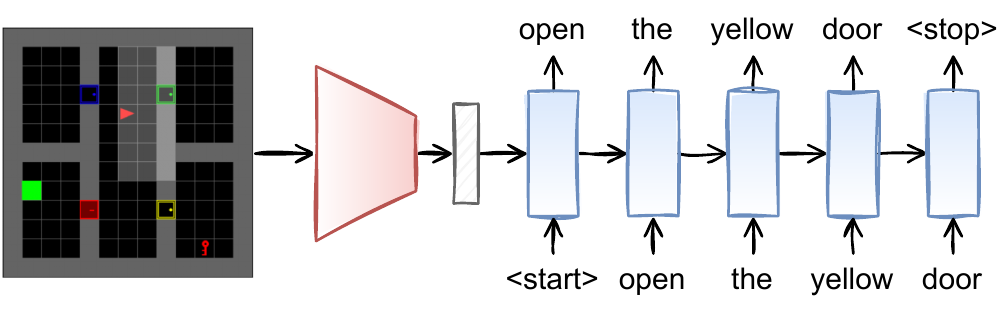}}\quad
  \hspace{2em}%
  \subfloat{\includegraphics[scale=0.70]{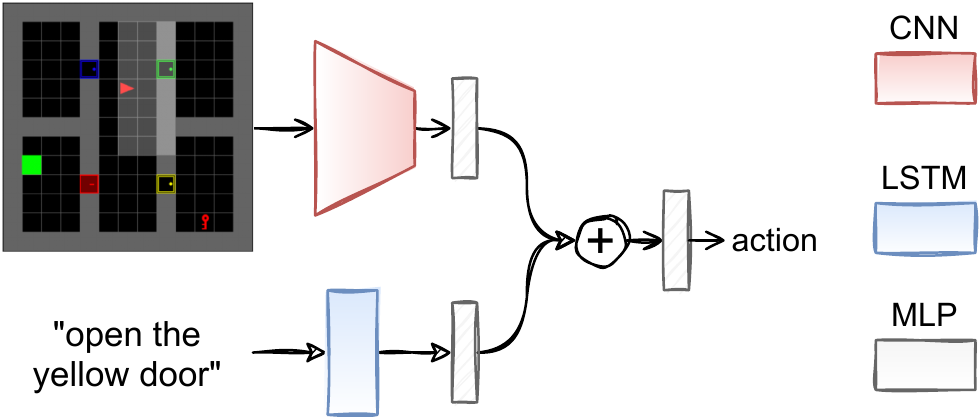}}\quad
  \hspace{1em}%
  \caption{(Left) Sub-goal instruction generator: It accepts an image observation which is processed using a CNN to output an image embedding. This is then used as the input to the LSTM which generates a language instruction. (Right) Low-level Controller: It is a language conditioned policy where it accepts both an image embedding and a language instruction as input and output an action. The image is processed using a CNN and the language instruction using an LSTM. Both the embeddings are simply concatenated and passed through fully-connected layers to predict the action.}
\label{fig:hrl_arch2}
\end{figure*}

As show in Figure \ref{fig:hrl_arch}, our agent has a hierarchical structure where the high-level module observes the current state and issues a high-level sub-goal using language. The low-level controller is a language conditioned policy trained using reinforcement learning (RL) which accepts the language instruction and executes actions in the environment. The sub-goal generator is trained in a supervised way using a small number of expert demonstration data. Since collecting large amounts of expert data is challenging we test this module with varying amounts of data. 
This also means that the sub-goal instruction generator might not be always accurate in providing sub goals. During testing, we also employ a human to intervene and take over the job of this module and provide language instructions. 

We test this on the MiniGrid environment \cite{gym_minigrid} where the goal of the agent is to navigate a number of different rooms, using keys and doors and ultimately reach the exit. These kinds of long horizon tasks a very difficult to solve using standard RL and IL. And this is especially true when the reward is sparse. We show that leveraging the hierarchical structure of tasks as well has using language is beneficial both in terms of sample efficiency and interpretability.

\section{Related Work}

Reinforcement Learning agents can learn to solve complex sequential decision making tasks (i.e. MDPs) by interacting with the environment, collecting experience, and learning from that experience \citep{sutton1998introduction, mnih2015human}. Methods like DQN \cite{mnih2015human}, PPO \cite{schulman2017proximal}, Deep TAMER \cite{Warnell2018} have used deep networks to solve tasks with high dimensional state and action spaces. If we have access to expert human demonstrations, we can use imitation learning algorithms to learn policies my 'imitating' the expert actions \cite{argall2009survey} \cite{ross2011reduction}. However, a longstanding problem with RL and IL is the relatively poor sample efficiency during learning which can lead to generalization and scalability issues when trying to solve complex tasks. In our work we tackle this by dividing tasks into sub-tasks and train an agent with multiple levels of control. There are several approaches to learn hierarchical agents as seen in \cite{sutton1999between} and \cite{fruit2017exploration}. Our approach is more similar to \cite{andreas2017modular} which assumes access to sub-tasks but nothing about how to execute them and the right order.

Language can be used to specify plans, goals, and high level requirements to each other \cite{gopnik1987development}. We humans can learn to do tasks in new environments not only from demonstrations, but also from information encoded using language \cite{tsividis2017human}. Traditional Reinforcement Learning and Imitation Learning approaches do not really attempt to ground language and environment features. Although there are many non verbal ways to communicate, language is an effective widely used mode of communication and obviously has many advantages. In most current research, language is used in RL in two main ways, language conditioned RL and language-assisted RL \cite{macmahon2006walk} \cite{hermann2017grounded}. Methods developed for language conditional tasks are relevant for language-assisted RL as they both deal with the problem of grounding natural language sentences in the context of RL \cite{goyal2019using} \cite{bahdanau2018learning}. 

There has been some work in learning mappings between language and actions or behaviors. \cite{branavan2009reinforcement} shows a way to learn mapping between language instructions and action in reinforcement learning. They show this on simple game tutorials and troubleshooting environments. \cite{chen2011learning} present a system that learns a semantic parser for interpreting navigation instructions by simply observing the actions of human followers and recent work has shown the utility of natural language narrations for guiding RL policies to learn complex tasks such as StarCraft 2 \cite{Waytowich2019a,waytowich2019b}.

\begin{figure*}
  \centering
  \subfloat{\includegraphics[scale=0.30]{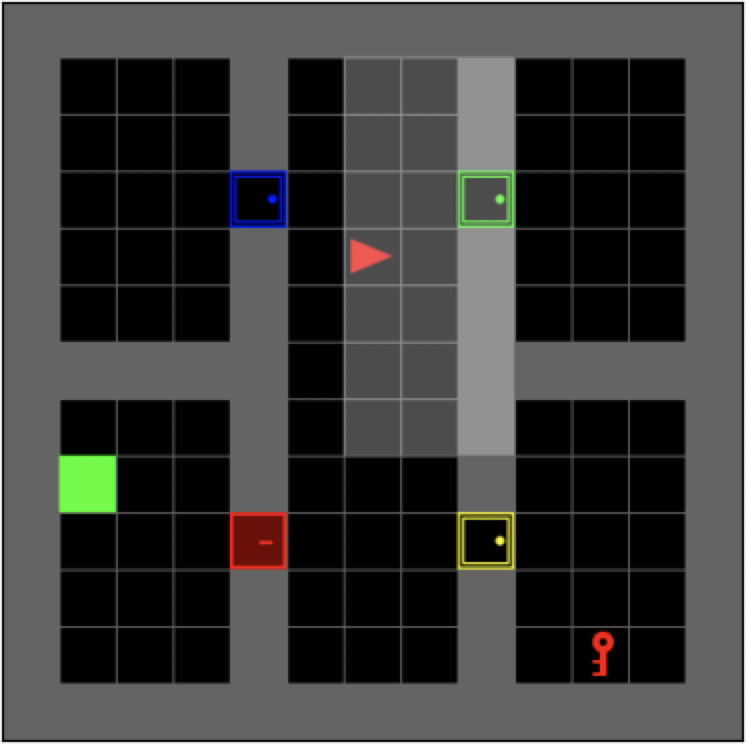}}\quad
  \hspace{3em}%
  \subfloat{\includegraphics[scale=0.30]{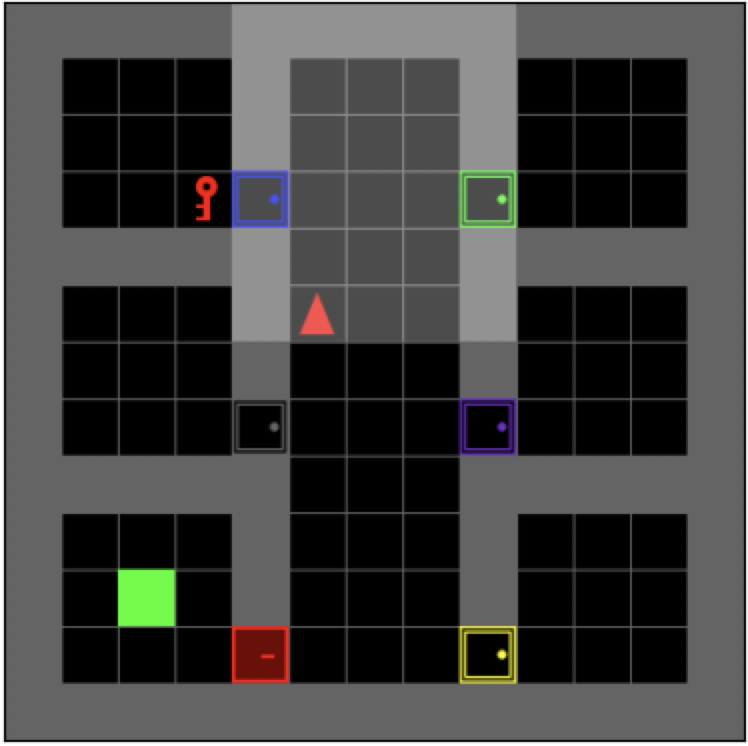}}\quad
  \caption{\textbf{MiniGrid Environment:} (Left) 4 Rooms The goal is for red agent to reach the green exit by navigating rooms and manipulating objects like keys and doors. In 4 rooms environment, the task is divided into 4 sub-tasks -- open the yellow door, pick up the red key, open the red door, go to goal. (Right) The 6 rooms environment is similar but has additional rooms and colors.}
\label{fig:env_desc}
\end{figure*}

\section{Methods}

We consider a framework where the agent consists of two modules, a high-level planner and a low-level controller. The high-level planner is responsible for providing high level sub-tasks or macro-actions. This can be either in the form of dense symbols using a vector or a more interpretable form like a language instruction. In our case, we represent this using language. The low-level controller takes the high-level macro action and learns to perform the sub-task.

Sub-tasks (or sub-goal) can be denoted by $g \, \epsilon \, G$ and low level actions as $a \, \epsilon \, A$. The state space can be denoted by $s \, \epsilon \, S$. Given this, the high level planner observes the state $s$ and chooses a sub-task $g$ which is in the form of a language instruction. This is then performed by the low-level controller by executing a series of actions $a$. After which the high-level controller picks a new sub-goal and the process repeats.

The goal now is to learn both the high level policy $\pi_g : S \rightarrow G$ and the low level level policies or sub-task policies $\pi_c : S \rightarrow A$ for each  $g \, \epsilon \, G$. Ultimately, we want the agent to achieve high rewards when we run both the high level and low level policies together. 

Our architecture is shown in Figure \ref{fig:hrl_arch} where we have two main modules, the sub-goal instruction generator which is the high-level planner and a low-level controller. The sub-goal instruction generator takes in the current state and generates a language instruction corresponding to a sub-task. The low-level controller then accepts this instruction and performs a series of primitive actions to to achieve the sub-goal. We assume that the horizon for the low level controller is $H_l$. For this work, we choose a fixed value $H_l = 10$. Which means that the low-level controller has 10 steps to complete sub-goals after which sub-goal instruction generator is used to get a new sub-goal. Future work can explore ways to detect termination conditions making this more robust. 

In order to train the sub-goal instruction generator we collect expert demonstrations along with high-level language instructions corresponding to the sub-tasks. This is set up as a supervised learning problem where the input to the model is a state and the the output is a language instruction. We train the low level controller separately and it is setup as a multi-task problem. In the end we get a language conditioned policy which can execute sub tasks.

\subsection{Data Collection}

We collect expert demonstrations using an expert policy(or human)  where we create a dataset of mappings between state and high-level language instruction. In our experiments, we were able to automate the process using an expert bot and we used a small, fixed grammar to generate language instructions. 

\subsection{Sub-goal Instruction Generator}

The sub-goal instruction generator is an encoder-decoder framework. Here, the encoder is the state encoder and the decoder is an LSTM which outputs a language instruction as shown in Figure \ref{fig:hrl_arch2}. At each time step, the sub-goal instruction generator receives a state which is encoded by a state encoder. In this case is a convolution neural network (CNN) which outputs a state embedding. The CNN has 3 layers of 16, 32 and 64 filter each with a kernel size of 2. The image embedding size is 512. This is then decoded by the LSTM decoder which has a an input size of 512 and hidden size of 1024. The target to the LSTM is the language instruction in the demonstration dataset. 

\subsection{Low-level Controller}

The low-level controller is a simple language conditioned policy which accepts a state and language instruction at each time step. Similar to the sub-goal instruction generator, the state is encoded using a CNN state encoder and the language is encoded using an LSTM. The architecture of both the CNN and LSTM is similar to the one described earlier. Both the state and language encoding are concatenated are passed  through 2 fully connected layers of size 64 and the final layer of size 7 outputs a low level action.

\begin{table}[t]
  \centering
    \begin{tabular}{l c c}
    \toprule
    \multicolumn{1}{l}{\textbf{No. of Demos}} & \textbf{TC \%}  & \textbf{Avg. HI} \\
     \toprule
    50  &   0.30    &   5.9  \\
    100 &    0.55   &    4.85  \\
    500 &    0.90   &   1.05   \\
    1000 &   0.95    &   0.5   \\
    \bottomrule
    \end{tabular}
  \label{tab1}
\end{table}%

\begin{table}[t]
  \centering
    \begin{tabular}{l c c}
    \toprule
   \multicolumn{1}{l}{\textbf{No. of Demos}} & \textbf{TC \%}  & \textbf{Avg. HI} \\
     \toprule
    50  &   0.15    &   7.7  \\
    100 &    0.30   &    6.1  \\
    250 &    0.65   &   5.05   \\
    500 &    0.75   &   3.85   \\
    1000 &   0.90    &   1.26   \\
    \bottomrule
    \end{tabular}%
    \caption{(Top) Results for 4 Rooms task, (Bottom) Results for 6 Rooms task. \textbf{TC \%} is task completion \% without any human intervention. We also employ a human expert to oversee the sub-goal instruction generation. We measure the average number of times a human had to intervene in order to correct the high-level language sub-goals. This is denoted by \textbf{Avg. HI}.  We show the ablation for this method with varying amounts of human demonstrations used to train the  }
  \label{tab2}
\end{table}%

\section{Experiments}

In this section, we will explain the experimental setup as well as the results. Our experiments are performed on the MiniGrid Environment by \cite{gym_minigrid} which is a simple grid world environment. The main goal of our experiments is to understand how well our hierarchical model can generate language instructions and learn to interpret them. We also show the ability for a human expert to intervene and provide language feedback whenever necessary.

We show some ablations on number of demonstrations and human interventions needed to complete the tasks. The tasks we test on are shown in Figure \ref{fig:env_desc}. The the final goal of the agent is to reach the green exit which is always in a locked room. The agent first has to find the key from by going into the right room, pick up the key, open the locked room and then reach the goal. We test on 2 versions of this task with 4 and 6 rooms. The positions of keys, goal, agent, and door colors are randomly initialized for each episode. Example instructions are shown below.   


\begin{table}[h]
  \centering
    \begin{tabular}{l}
    \toprule
    \multicolumn{1}{l}{\textbf{Sub-goal Instructions}}\\
     \toprule
    open the yellow door \\
    pick up the blue key  \\
    go to the goal \\
    \bottomrule
    \end{tabular}%
  \label{tab3}%
\end{table}%

We first train the low level controller using RL in a multi-task setup as the first step. The reward function is design in such a way to provide a binary positive reward on completion of the individual sub-goal. We sample random sub-goals in the form of language instructions at each episode and give a positive reward on successful completion of the sub task. We design a slightly modified version of the environment for this step where any sub-goal can be achieved independently. We use \cite{schulman2017proximal} with the default settings. Another potential option is to use the available demonstrations to warm start the agent using imitation learning (IL). We leave that as future work.

As discussed in earlier sections, we train the sub-goal instruction generator using demonstration data. In Table 1, we show the performance of the agent when we run both the policies together. We report the independent task completion percentage for both the tasks with different number of demonstrations. This is denoted by TC \%. We run the evaluation again with an expert human who oversees the agent and issues correct language commands whenever the high level module fails. Here we report the avg number of human interventions/corrections per episode required in order to complete the task. For all the cases the number show an average over 50 runs. Table 2 shows comparison with standard flat RL baselines with both sparse and dense rewards with roughly same amount of training budget in terms of time steps.

\begin{table}[t]
  \centering
    \begin{tabular}{l c c}
    \toprule
    \multicolumn{1}{l}{\textbf{Method}} & \textbf{4 Rooms TC \%} & \textbf{6 Rooms TC \%}  \\
     \toprule
    RL (Sparse) &    30   &   15   \\
    RL (Dense) &    70   &   53  \\
    Our (500D)  &   \textbf{90}    &   \textbf{75}  \\
    Our (1000D) &    \textbf{95}   &    \textbf{90}  \\
    \bottomrule
    \end{tabular}
    \caption{Comparison with standard RL methods with both dense and sparse rewards. All baseline RL models were trained for 4M steps. Our method was 3M steps but with small number of some expert demonstration data. 500D and 1000D denote no. of demos.}
  \label{results2}
\end{table}%

\section{Conclusion}

In this work, we propose a hierarchical framework where high level sub-goals are generated in language and these are executed by a low level controller. We show that this method performs better than standard RL and is much more sample efficient. It also has an added benefit of interpretability where an expert can observe the language commands being issued by the high level planner and intervene if necessary. The human can then issue the correct sub-goal in language for which is much more natural and intuitive. Future work includes testing this method on more complex tasks with high-dimensional and continuous state and action spaces.

\bibliography{rl.bib}

\end{document}